\documentclass[conference]{IEEEtran}
\IEEEoverridecommandlockouts
\usepackage{cite}
\usepackage{amsmath,amssymb,amsfonts}
\usepackage{algorithmic}
\usepackage{graphicx}
\usepackage{textcomp}
\usepackage{xcolor}
\def\BibTeX{{\rm B\kern-.05em{\sc i\kern-.025em b}\kern-.08em
    T\kern-.1667em\lower.7ex\hbox{E}\kern-.125emX}}
\begin{document}

\title{Evolving Agents for the Hanabi 2018 CIG Competition\\
}

\author{\IEEEauthorblockN{Rodrigo Canaan}
\IEEEauthorblockA{
\textit{New York University}\\
New York, US \\
rodrigo.canaan@nyu.edu}
\\
\IEEEauthorblockN{Julian Togelius}
\IEEEauthorblockA{
\textit{New York University}\\
New York, US \\
julian.togelius@nyu.edu}
\and
\IEEEauthorblockN{Haotian Shen}
\IEEEauthorblockA{
\textit{New York University}\\
New York, US \\
hs2685@nyu.edu}
\\
\IEEEauthorblockN{Andy Nealen}
\IEEEauthorblockA{
\textit{New York University}\\
New York, US \\
nealen@nyu.edu}
\and
\IEEEauthorblockN{Ruben Torrado}
\IEEEauthorblockA{
\textit{New York University}\\
New York, US \\
rrt264@nyu.edu}
\\
\IEEEauthorblockN{Stefan Menzel}
\IEEEauthorblockA{
\textit{Honda Research Institute Europe GmbH}\\
Offenbach, Germany \\
stefan.menzel@honda-ri.de}
}


\maketitle

\begin{abstract}
Hanabi is a cooperative card game with hidden information that has won important awards in the industry and received some recent academic attention. A two-track competition of agents for the game will take place in the 2018 CIG conference. In this paper, we develop a genetic algorithm that builds rule-based agents by determining the best sequence of rules from a fixed rule set to use as strategy. In three separate experiments, we remove human assumptions regarding the ordering of rules, add new, more expressive rules to the rule set and independently evolve agents specialized at specific game sizes. As result, we achieve scores superior to previously published research for the mirror and mixed evaluation of agents.
\end{abstract}


\section{Introduction}

Game-playing agents have a long tradition of serving as benchmarks for AI research. However, traditionally most of the focus has been on competitive, perfect information games, such as Checkers~\cite{schaeffer1996chinook}, Chess~\cite{campbell2002deep} and Go~\cite{AlphaGo}. Cooperative games with imperfect information provide an interesting research topic not only due to the added challenges posed to researchers, but also because many modern industrial and commercial applications can be characterized as examples of cooperation between humans and machines in order to achieve a mutual goal in an uncertain environment. In this paper, we address a particularly interesting cooperative game with partial information: Hanabi~\cite{hanabibgg}.

Hanabi is a cooperative card game designed by Antoine Bauza released in 2010 where 2 to 5 players play with their hands facing outwards, so that only the content of the other players' hand is seen. They can only communicate through a limited number of hints, which allow a player to point to all cards of a chosen value or color in another player's hand. The objective is to build one stack for each of the five colors by playing cards in ascending value order (from 1 to 5). By discarding a card or completing a stack, one hint token is replenished to the group. The group wins if all stacks are complete (corresponding to a score of 25). The group loses a life when a card is played out of order. If all lives are used, the game ends immediately. If the draw deck runs out, the game ends after one last turn for each player. In either case, the game does not count as a win, but a partial score is computed by adding the size of each stack (or equivalently, counting the number of played cards). A more rigorous explanation of the rules can be seen in ~\cite{osawa2015solving}. 


Each card has a value (also called rank or number) in the range of \{1-5\}, and a color (or suit) out of \{B, R, Y, W, G\}. From now on, a card will be referred to as (CV) where C is its color and V is its value. For example (R2) denotes a red card with value 2. If only partial information is known, we represent only the color or the value. For example, (Y) is any yellow card and (5)  is any card of value 5. Keep in mind that by telling a player of a color or value, all cards not pointed are known to be of some different color or value. We call this knowledge ``negative information". Although the agents discussed in this work are able to reason with negative information, we do not include it in our notation for simplicity.

We will denote the number of players in the game as \#players or game size. The starting number of cards in each player's hand depends on \#players: 5 cards for a game size of 2 or 3, and 4 cards for a game size of 4 or 5 players.


The game was well received by the tabletop games community, winning the \textit{Spiel des Jahres} award in 2013~\cite{spiel}, and has also received attention by game AI researchers for being a challenging cooperative problem with hidden information and a limited, but well-defined communication channel.

Because both the number of hint tokens and the number of copies of each card in the deck are limited, some challenging decisions that often arise in a Hanabi match are wether to play a card with only partially known information (and risk it being unplayable) or wait for more information (using up a hint token), whether to discard a card with partial information (and risk it being the last copy of its kind in the deck) and, from the other side of the table, whether to allow other players to make these risky decisions or to use a hint token to clarify the situation. Belief about how other players operate can be a key factor in making such decisions.

In 2018, the game will feature in a two-track competition~\cite{competition} at the CIG conference. In the Mixed-Track, each agent will play with unknown agents, while in the Mirror-Track, each agent will play with copies of itself and so the behavior of other players can be assumed for the most part (except possibly for stochastic behavior and hidden information). The competition page also provides the framework used in this research, which includes a sample rule-based agent, and a vast library of rules.
It also features implementations of other sample agents, such as a random agent and a MCTS agent that can optionally be equipped with a perfect model of its playing partners.

In this paper, we review existing literature on Hanabi-playing agents and present our evolutionary approach to evolving rule-based Hanabi agents. We plan to submit our best agents to both tracks of the competition.

\section{RELATED WORK}

Optimal play of (generalized) Hanabi has been proven to be an NP-Complete problem by Baffier \textit{et al}.~\cite{baffier2016hanabi}, even if we remove all hidden information. Due to this complexity, most research in Hanabi-playing agents~\cite{osawa2015solving,van2016aspects,eger2017intentional,walton2017evaluating} focuses on one of two methods (or a combination of both):
\begin{itemize}
\item  Rule-based  agents, which go through an ordered list of heuristics (such as play a card that is known to be playable, discard a random card, hint a playable card, etc.) and play the first one that is applicable.
\item Search-based agents, which use a model of the other players' behavior to search for the action that leads to best results, such as Monte-Carlo Tree Search (MCTS)~\cite{browne2012survey}, or to search for previous states that are consistent with the hints received by another player.
\end{itemize}

Osawa's~\cite{osawa2015solving} best performing agent enhances a sequence of rules with a search of all possible previous states that are consistent with the other player's last action.

Van den Berg \textit{et al}.~\cite{van2016aspects} optimize a rule-based agent by exhaustively searching 48 possible agents obtained by selecting one of 4 possible hint heuristics, 4 discard heuristics and 3 thresholds for playability of a card. The order of application of the heuristics was pre-defined. Each strategy was evaluated by their average score in simulated mirrored play. They also implement a MCTS agent for the game, which does not perform better than their best rule-based agent.

Eger \textit{et al}.~\cite{eger2017intentional} propose an intentional rule-based agent that simulates the best hint to give, assuming a model of the other player. They also validated their agent by playing with human players (achieving lower score than in mirrored play).

All agents described so far (other than Eger's human evaluation) have in common that they assume they will be playing with agents following a similar strategy. Walton-Rivers \textit{et al}.~\cite{walton2017evaluating} address the problem of  playing with a diverse population of agents with different strategies. They use several rule-based agents, including reimplementations of Osawa and Van den Berg, along with  MCTS agents which either receive a model of the other agents' behavior or assume random play. They evaluate their agents by pairing them with a fixed set of baseline agents, and their best performing agent is called Piers, which achieves a score of 11.8 with that specific test pool. Because this paper was written by the author of the Java framework being used and by the organizer of the CIG competition, it is central to our work and will sometimes be referred to as ``the original paper/article''.

The best existing agents for mirrored-play are those by Cox \textit{et al}.~\cite{cox2015make}. They treat the game as a hat-guessing game. Each hand is assigned a value by a publicly-known algorithm (e.g. 1 means "first card is playable", 2 means "first card is discard-able") and each possible hint is encoded as a number that is interpreted to be the sum (mod number of players) of the values of all other players' hands. This means they manage to give a clear instruction (play or discard a specific card) to every player with a single hint. Their version of hat-guessing agent can play only games with 5 players.

Bouzy~\cite{bouzy2017playing} generalizes hat-guessing players for game sizes 2-5. He also proposes a rule-based agent called Confidence that achieves scores of 18.16 across all game sizes. He the uses both his hat agents and the Confidence agents as evaluators for and Tree Search agents, achieving even higher scores. However, their tree search agents reportedly use 10 seconds per move on average, which makes them unsuitable for the competition as moves are expected to be returned in 40ms.

Table \ref{table:existing-results} shows a summary of the best agents in literature, with their evaluation mode (mirror, mixed or human play) and techniques used.  These numbers give us a benchmark to measure our results against. We consider results above 18.16 for mirror play and above 11.18 for mixed play (with the same test pool as in ~\cite{walton2017evaluating}) to be an improvement for our purpose, as the hat agents require a fixed convention that is unsuitable for mixed or human play, and the Tree Search agent exceeds our time budget per move.

Note that some agents discussed in this section are specialized for a specific game size, and others are able to play with a number of cards per player different than the official rules. Column \textbf{\#Players} of table \ref{table:existing-results} denotes what games sizes the agent is capable of playing in. We consider only the reported scores with the standard number of cards per player and, for agents able of playing multiple game sizes, we average the score across all game sizes.

\begin{table}[]
\centering
\caption{Existing results}
\label{table:existing-results}
\begin{tabular}{|p{2cm}|l|p{1.2cm}|l|p{2cm}|}
\hline
\textbf{Author}       & \textbf{Score} & \textbf{Evaluation} & \textbf{\#Players} & \textbf{Technique}       \\ \hline
Osawa                 & 15.85          & Mirror              & 2&  Rule-based / Search      \\ \hline
Van den Bergh         & 15.4           & Mirror              &3&  Rule-based               \\ \hline
Eger                  & 17.1           & Mirror              & 2& Rule-based / Intentional \\ \hline
Eger                  & 14.99          & Human Play          & 2& Rule-based / Intentional \\ \hline
Cox                   & 24.68          & Mirror              & 5& Hat-Guessing             \\ \hline
Bouzy - Confidence                   & 18.16         & Mirror              & 2-5& Rule-Based            \\ \hline
Bouzy - Tree Search with Confidence                   & 20.22         & Mirror              & 2-5& Tree Search            \\ \hline
Walton-Rivers (Piers) & 11.18          & Mixed               & 2-5& Rule-based               \\ \hline
\end{tabular}
\end{table}

\section{METHODOLOGY} \label{section:methodology}

One of the main major gaps in current research is that  rule-based  agents are either hand-crafted by human experts or result of exhaustive search in a narrow search space, such as Van den Bergh \textit{et al}.~\cite{van2016aspects}, who specifies a sequence of rules to be applied and then searches for the best of 48 possible selections of parameters. We propose to instead use an evolutionary algorithm to determine the rules, their order and parameters with no further human assumptions.

In this section, we rigorously define what we mean by rule-based agents and rules, how mirror and mixed evaluation are performed and how the genetic algorithm works, in order to attempt to build better agents than the human-crafted ones.

\subsection{Definitions}

We define a \textbf{rule-based agent} as one that scans a list of rules ordered by priority, and immediately plays the action implied by the first applicable rule. A \textbf{rule} is defined as a function that takes a game state and the current active player and returns either a legal action, if the rule is applicable, or null if the rule is not applicable. For example, a rule that tells a player a random piece of information of a playable card would fail to return a value if no other player has a playable card or if no hint tokens are available. Otherwise, it would return the action to hint the color or value of the playable card to the player who holds that card. In trying to apply a rule, only information that is available to the active player can be used. 

A common pattern in all human-created agents and most of the successful evolved agents is to execute rules roughly in the following order:
\begin{itemize}
\item Play a ``good'' card
\item Give a hint to a player about a ``good'' card
\item Discard a ``useless'' card
\item Tell a player about a ``useless'' card so they can discard it
\end{itemize}
The specific implementation of a rule defines the meaning of what constitutes a good or useless card, how to break ties between two or more cards, which player to give a hint to and which hint to give if multiple pieces of information (color and value) are missing. Usually, a ``good card'' is a card that is (or has a high probability of being) immediately playable, whereas a ``useless'' card is a card that is never going to be playable again (because the stack of that color is at a higher number than the card, or all prerequisites were accidentally discarded). Some rules also care whether a card is ``necessary'', meaning that discarding it would prevent players from ever  completing a stack.

Table ~\ref{table:ruleset} in the appendix gives a short description of each rule used. Some rules were already natively implemented in the framework. These are classified as categories 1 and 3 in the table. Category 1 is for rules very similar to the ones described in ~\cite{walton2017evaluating}, focusing on the probability (from its owner's perspective) that a card is playable , or discard-able. They were already implemented as classes in the framework. Category 2 is for rules that also appear in ~\cite{walton2017evaluating}, but are not natively available as classes in the framework and had to be implemented separately by using the framework' functionality for conditional rules. We refer to the original article for an in-depth explanation of those rules. Category 3 is for native rules in the framework but do not correspond to rules in ~\cite{walton2017evaluating} and attempt to implement specific human-created strategies discussed in a strategy forum~\footnote{https://boardgamegeek.com/thread/1309490/finesse-bluff-reverse-finesse-explained  and https://www.boardgamegeek.com/article/23427635\#23427635\\ Access:05/15/2018}. More detail on those rules can be found in the framework documentation. 

Rules in category 4 were implemented by us for the purpose of this work and are described in detail in section ~\ref{section:experiments}.

\subsection{Mirror and mixed evaluation} \label{sub:mirror-mixed}

We propose the use of a genetic algorithm to determine the rules as well as their number and order to play Hanabi for different numbers of players. Agents can be evolved using two kinds of evaluation as fitness function: mirror-play performance and mixed-play performance. In \textbf{mirror-play}, an agent is paired with copies of itself, and plays $n$ games on each of  the 4 game sizes. The fitness of an agent is the average score in all $4n$ games.

For \textbf{mixed-play}, we use the same test pool of seven agents as used by ~\cite{walton2017evaluating}, consisting of the following seven agents: IGGI, Internal, Outer, Legal Random, Van den Bergh, Flawed, and Piers. While we do not know the exact test pool that will be used in the competition, this set of agents was used by the authors of the competition in their previous experiments, and it contains both well-performing agents (such as Piers) and intentionally poor-performing agents (such as Legal Random and Flawed). It also contains some agents that play in a totally deterministic way and others that behave stochastically. For this reason, we expect performance with this test set to be good indicator of performance in the overall mixed competition.

For an explanation of how these agents work, see ~\cite{walton2017evaluating}. For each game size $s\in\{2,3,4,5\}$, the agent being evaluated is placed in a random starting position and plays with $s-1$ copies of the same agent from the test pool. Each player plays $n$ games for each of the seven pairings for each of the four possible game sizes, for a total of $28n$ games per generation per agent. Once again, the fitness of an agent corresponds to the average score of all games. Note that we do not know the actual agents that will be used as test pool in the mixed track of the competition, so we use performance in the test pool of the original article as an indicator of how well our agent plays with unknown agents.

To reduce the effects of randomness, we use the same random seeds to determine the starting position and starting deck configuration for all players in a population.

\subsection{Genetic Algorithm}
\begin{table}[]
\centering
\caption{Parameters of the genetic algorithm}
\label{table: GA}
\begin{tabular}{|l|l|}
\hline
\textbf{Parameter}        & \textbf{Value} \\ \hline
Population size (p)       & 200            \\ \hline
Chromosome size (s)       & 50 or 72       \\ \hline
Mutation rate (m)         & 0.1            \\ \hline
Crossover rate (c)        & 0.9            \\ \hline
Elitism count (e)         & 20             \\ \hline
Tournament size (t)       & 5              \\ \hline
Number of generations (G) & 500            \\ \hline
Games per generation (n)  & 20             \\ \hline
\end{tabular}
\end{table}

The main objective of our genetic algorithm is to determine a good ordering of the existing rules in the rule set. This kind of problem could be categorized
as a combinatorial problem. Each rule could be represented as an integer number which does not repeat within the chromosome. 

The use of Genetic Algorithm to solve combinatorial problems has been studied deeply in the literature, \cite{anderson1994}, \cite{moraglio2006}. We use the operators of swap mutation (with probability $m$) and ordered crossover (with probability $c$) , which maintain the constraints that each rule is selected only once per chromosome. Individuals are chosen for crossover using tournament selection ~\cite{miller1995genetic} with tournament size $t$. An elitism count of $e$ is enforced.

Each of the $p$ chromosomes is initialized as a random ordering of $s$ distinct rules in the rule set (we used a chromosome size equal to the size of the rule set in our experiments: 50 for the original rule set and 72 for the new rule set, although smaller chromosome sizes were tested with no significant impact in score or algorithm performance). For each of the $G$ generations, the fitness function is the result of playing $n$ games for each game size and each pairing in mirror or mixed evaluation as described in section ~\ref{section:methodology}.

Table~\ref{table: GA} summarizes the parameters of our genetic algorithm.

\section{EXPERIMENTS AND RESULTS} \label{section:experiments}

In this section, we describe the experiments performed. All scores reported are obtained by averaging the score of an equal number of games with each possible game size (2 to 5 players), except for the specialized agents, who either play only two-player games or only 3-5 player games. All games are played with the standard hand sizes: 5 cards per player with 2 or 3 players and 4 cards per player with 4 or 5 players.

\subsection{Validation results}

Our first step was to reproduce the agents described in ~\cite{walton2017evaluating}. We implemented the seven rule-based agents described in subsection~\ref{sub:mirror-mixed}. Note that the original article implements other agents, such as MCTS variations, but the well-performing ones do not outperform the best rule-based agent (Piers) and require a precise model of the other players, which is not available to us.

In the original paper, Walton-Rivers uses those seven agents as his test pool, and six of them (all except internal) are evaluated against this test pool. Our validation experiment consists of pairing the six agents with the seven agents in the pool. Table~\ref{table:validation} shows the results of our validation, with most agents getting very similar results to the score reported by the original author, leading us to believe our implementation of the test pool is valid, which is an important initial step for the mixed evaluation experiments below.

\begin{table}[]
\centering
\caption{Validation Results}
\label{table:validation}
\begin{tabular}{|l|l|l|l|}
\hline
\textbf{Agent} & \textbf{Their Score} & \textbf{Our Score} & \textbf{s.e.m} \\ \hline
IGGI           & 10.96                & 10.98              & 0.06           \\ \hline
Outer          & 10.2                 & 9.70               & 0.05           \\ \hline
Legal Random   & 4.59                 & 4.52               & 0.04           \\ \hline
Van Den Bergh  & 10.88                & 11.02              & 0.06           \\ \hline
Flawed         & 5.02                 & 4.46               & 0.04           \\ \hline
Piers          & 11.18                & 11.28              & 0.06           \\ \hline
\end{tabular}
\\Note: Number of games = 4*7*400 = 11200 per agent evaluated
The standard error of the mean (s.e.m.) reported here corresponds to the error in our experiment, not theirs.
\end{table}

\subsection{Evolution using the existing rule set}

For this experiment, we ran the evolutionary algorithm (in both mixed and mirror mode) using only the rules native to the framework, plus the conditional rules necessary to run the agents in the test pool. These are marked as categories 1, 2 and 3 in table~\ref{table:ruleset}. Since most of there rules (except category 3) correspond to rules described in~\cite{walton2017evaluating}, our objective for this experiment was to verify if, by throwing away any human assumptions of which order rules should be applied, but using a very similar rule set as the original article, we could get agents that outperformed the hand-crafted agents, in particular agent Piers, which is the best agent described in ~\cite{walton2017evaluating}.

\begin{figure}
\centering
\includegraphics[width=0.5\textwidth]{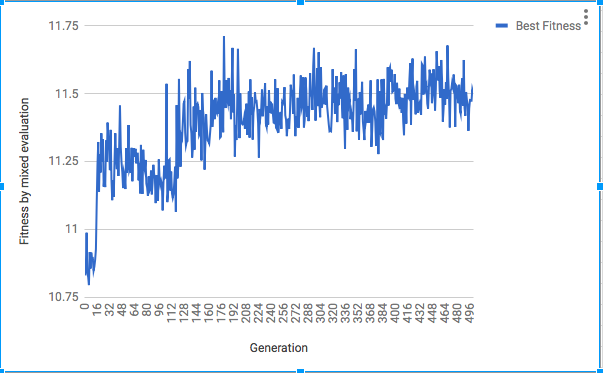}
\caption{\label{fig:fitness} An illustration of how fitness of the best agent varied per generation in one of our evolutionary runs for mixed mode with old rules}
\end{figure}

Because fitness is calculated by a number of simulations using random seeds, the fitness of an agent can fluctuate with each generation, even if the chromosome is unchanged (due to elitism). See figure~\ref{fig:fitness} as an illustration of this fact.  For that reason, after running the algorithm for 500 generations we took the agents corresponding to the 10 best performing chromosomes and ran a second round of simulations. The best mirror and mixed agent in this second round of simulations are referred to as MirrorOld and MixedOld in tables \ref{table:mirror} and \ref{table:mixed}, which also shows their performance in all game sizes, the number of games played per agent and the standard error of the mean. We manage to beat the baseline on both mirror (from 18.16 to 18.38) and mixed (11.18 to 11.45) modes.

Table~\ref{table:mixed-old} shows the best evolved chromosome for mixed play using only these ``old'' rules. Note that the first two rules are redundant, as if a card is more than 80\% likely to be played, it is also more than 60\% likely to be playable, but would not be redundant if there was another rule between them.

\begin{table}[]
\centering
\caption{Chromosome for mixed play using only ``old'' rules}
\label{table:mixed-old}
\begin{tabular}{|l|l|}
\hline
\textbf{Rule name}                                \\ \hline
IF(Lives\textgreater{}1) PlayProbablySafeCard(0.8) \\ \hline
IF(Lives\textgreater{}1)PlayProbablySafeCard(0.6)  \\ \hline
TellAnyoneAboutUsefulCard                         \\ \hline
PlaySafeCard                                      \\ \hline
DiscardProbablyUselessCard(0.4)                   \\ \hline
DiscardUnidentifiedCard                          \\ \hline
DiscardOldestFirst                                \\ \hline
TellDispensable                                  \\ \hline
TellRandomly                                      \\ \hline
\end{tabular}
\end{table}

For space considerations, we do not show here the best chromosomes for the other categories, but we will make them publicly available in our repository after the competition\footnote{https://github.com/rubenrtorrado/hanabi}.

\begin{table*}[]
\centering
\caption{Results for mirror - Number of games = 4*2000 = 8000 per agent evaluated. Baseline is 18.16}
\label{table:mirror}
\begin{tabular}{|l|l|l|l|l|l|l|}
\hline
\textbf{Agent name} & \textbf{avg}&\textbf{2P} & \textbf{3P} & \textbf{4P}& \textbf{5P}&  \textbf{s.e.m} \\ \hline
MirrorOld      & 18.38    & 19.35                & 18.95 & 18.22  & 17.02    & 0.02           \\ \hline
MirrorNew      & 19.07    & 19.61                & 19.68 & 19.11  & 17.87  
  & 0.02           \\ \hline
MirrorSituational     & 19.32     & 20.07                & 19.58 & 19.36  & 18.29    & 0.02           \\ \hline
\end{tabular}
\centering{}
\end{table*}

\begin{table*}[]
\centering
\caption{Results for Mixed - Number of games = 4*7*2000 = 56000 per agent evaluated. Baseline is 11.18}
\label{table:mixed}
\begin{tabular}{|l|l|l|l|l|l|l|}
\hline
\textbf{Agent name} & \textbf{avg}& \textbf{2P} & \textbf{3P} & \textbf{4P}& \textbf{5P}& \textbf{s.e.m} \\ \hline
MixedOld     & 11.45      & 12.13               & 12.05 & 11.08  & 10.55   & 0.02           \\ \hline
MixedNew    & 11.50      & 12.11                & 12.19 & 11.15  & 10.56  
  & 0.02           \\ \hline
MixedSituational     & 11.65     & 12.38                 & 12.29 & 11.22  & 10.72    & 0.02           \\ \hline
\end{tabular}
\end{table*}

\subsection{New rules}

After establishing that the existing rule set allows us to build better agents than our baseline, we attempted to extend the rule set, searching for rules that implement behaviors that are not covered by the existing rules. The first rule we added was \textbf{PlayJustHinted}. This rule will attempt to play a card that was just hinted by another player, accounting for their likely intentions. For example, if a player points to a card that you just drew telling it is a (5), when a (4) has just been played, it is likely that the hint was motivated by the fact that it is the correct color. PlayJustHinted scans the event history for all hints received since our last action, and attempts to play the most likely playable card among those that were pointed at by a hint. Optionally, we can choose to play a card only if it was pointed as a standalone hint (a hint that tells of no other cards), only if it is also our most recently drawn card, only if the probability of it being playable being $>p$, only if the number of lives is $>n$ or any combination of the above criteria. In the rule set, we included all combinations of probabilities p in \{0, 0.2, 0.4, 0.6, 0.8\} and lives n in \{0, 1\} with and without the restriction of standalone card or newest card .

The other rules we added attempt to give hints in a clearer, less ambiguous way to our partners. Sometimes, out of multiple playable cards, we need to decide which to hint and whether to tell color or value. Given that many successful agents take risks in playing cards that are not 100\% sure to be playable, it is important not to give a hint that has the side effect of making a player believe an actually unplayable card to be playable. For example, a hint that points to three cards with value (2), when only one (1) has been played, leaves the player guessing which of them is the correct color. It might have been better to tell the color of the correct (2), especially if it happens to be the only card on its color. Conversely, if all (1)s had already been played, it is probably better to tell about all (2)s than hinting each color individually. The new rule \textbf{TellUnambiguous1} attempts to give information about a playable card by either maximizing the number of playable cards pointed to or minimizing the number of unplayable cards pointing to.

\textbf{TellUnambiguous2}, instead of looking at the number of pointed cards, calculates, from the other player’s perspective, the probability they would assign to each of their cards being playable. We do this for each possible hint. Each playable card is rewarded with a certain weight $w1 * p$, where $p$ is the probability that it is playable from the other player's perspective. Each unplayable card is penalized with a factor of $w2*p$. We select the hint that maximizes the value of a player’s hand in this way.

As before, we ran the evolutionary algorithm for 500 generations, then ran a secondary evaluation of the 10 best chromosomes. The resulting best agent is referred to as MirrorNew and MixedNew in tables \ref{table:mirror} and \ref{table:mixed}. We improve the results compared to the agents evolved from the ``old'' rule set from 18.38 to 19.07 (mirror) and 11.45 to 11.65. We also noticed that our new rules appeared at high frequency at the start of the successful chromosomes. In particular, often many variations of PlayJustHinted coexisted at the start of the same chromosome. TellUnambiguous1 was often the highest priority ``Tell'' rule, while TellUnambigous2 was less successful.

\subsection{Combining specialized agents}

At this point, all of our agents are evolved by playing with game sizes from 2 to 5. We theorized that the strategy for playing when there are two players can be different from when there are five players. To prove this, we evolved a set of agents specialized in 2-player games and another set specialized in 3 or more-player games. After running a secondary evaluation on the two sets of specialized chromosomes, and finding the best specialized agent for game size of 2 and the best for game size of 3+, we built a situational agent that is a combination of both. It will check the number of players, then change its behavior to the one most suited for that game size.

We then ran an additional evaluation of the situational agent, which increased our scores from 19.07 to 19.32 (mirror) and from 11.50 to 11.65 (mixed), as shown in tables \ref{table:mirror} and \ref{table:mixed}

\section{Analysis of the evolved chromosomes}

In this section, we look at the composition of the top 10 evolved chromosomes in each experiment in order to attempt to identify which rules were most successful and any discernible patterns that emerge in different experiments that could help explain why some strategies might work better for mirror or mixed play, or for specific game sizes. While we provide the frequency in which we observe some features in our most evolved chromosomes, we do not claim any statistical rigor to this analysis. Our findings serve only as illustration of our results and determining their validity could serve as a future research question.

For the purpose of this section, we analyze only the runs including the new rules. We picked the top 10 chromosomes from MirrorNew and MixedNew (unspecialized), the top 10 chromosomes specialized at 2 player games in MirrorSituational and MixedSituational and the top 10 specialized crhomosomes for games with 3 or more players in MirrorSituational and Mixed situational.

First, we examine the success of our new rules. In table~\ref{table:newrules}, column (I), we see how often some variation of \textbf{PlayJustHinted} was selected as the very first rule in a chromosome. In total, this happens 24 out of 30 times in mirror mode, but only 13 out of 30 times in mixed mode. This could be due to the fact that agents such as Legal Random and Flawed in the test pool do not favor giving playable hints over giving random hints. Thus, in mixed mode, assuming a card is playable just because it was hinted is a less reliable strategy.

Because most agents have a rule for playing a card as highest priority, TellUnambiguous was not as often chosen as first rule of the chromosome. Nevertheless, it was very often chosen as first among the tell rules (rules that result in giving a hint). In column (II) of table~\ref{table:newrules}, we see how often this happens for each game mode. Overall, it was the preferred tell rule 19 out of 30 times in mirror mode, and only 7 out of 30 times in mixed mode, most often in the variation TellUnambiguous1. This rule attempts to maximize the number of playable cards pointed to with each hint, or minimize the number of unplayable cards. As such, it pairs extremely well with PlayJustHinted, which is less popular in mixed mode and could explain the relative lack of success of TellUnambiguous in this scenario. A popular tell rule for mixed mode was CompleteTellUsefulCard, which gives complete information about a useful card and could be very effective with the agents in the test pool that require complete information for playing a card.

The last part of our analysis is regarding the difference in strategy from 2 player games to games with 3 or more players. Table~\ref{table:gamesize} shows how often a play or tell rule was selected as first of its chromosome for the specialized 2 player or 3+ player games. In games with many players, playing a card nearly always takes precedence over giving a hint, with 19 out of the 20 chromosomes following this pattern. In 2 player games, there is a fairly even split between agents that prioritize playing over telling (12 out of 20) and telling over playing (8 out of 20). We suspect that in 2 player games, it is often correct to hold a card you know is playable in order to give a hint to your partner, to avoid them accidentally discard a useful card, for example. In games with more players, players who already know a playable card should probably give higher priority to playing it, and leave the task of giving hints to other players. By choosing to give a hint instead of playing your card, you would consume a hint that could better be utilized by a player who knows little of their own hand (and would be forced to discard if they got to their turn without an available hint token).

\begin{table}[]
\centering
\caption{Prevalence of new rules by experiment}
\label{table:newrules}
\begin{tabular}{|l|p{2cm}|p{2cm}|}
\hline
\textbf{Game Type} & \textbf{PlayJustHinted as first in chromosome (I)} & \textbf{TellUnambiguous as first tell (II)} \\ \hline
Mirror 2 players         & 5/10                                           & 7/10                                   \\ \hline
Mirror 3+ players          & 10/10                                          & 10/10                                  \\ \hline
Mirror unspecialized  & 9/10                                           & 2/10                                   \\ \hline
\textbf{Mirror overall}  & \textbf{24/30 }                                          & \textbf{19/30 }                                  \\ \hline 
Mixed 2 players           & 4/10                                           & 6/10                                   \\ \hline
Mixed 3+ players         & 4/10                                           & 1/10                                   \\ \hline
Mixed unspecialized  & 5/10                                           & 0/10                                   \\ \hline
\textbf{Mixed overall}   &\textbf{13/30 }                                          & \textbf{7/30 }                                 \\ \hline
\end{tabular}
\end{table}

\begin{table}[]
\centering
\caption{Priority of Play and Tell rules by game size}
\label{table:gamesize}
\begin{tabular}{|l|l|l|}
\hline
\textbf{Game Type} & \textbf{Play before Tell} & \textbf{Tell before Play} \\ \hline
Mirror 2 players         & 7/10                      & 3/10                      \\ \hline
Mixed 2 players           & 5/10                      & 5/10                      \\ \hline
\textbf{2 players overall}            & \textbf{12/20}                      & \textbf{8/20}                     \\ \hline
Mirror 3+ players         & 10/10                     & 0/10                      \\ \hline
Mixed 3+   player       & 9/10                      & 1/10                      \\ \hline
\textbf{3+ players overall}            & \textbf{19/20}                      & \textbf{1/20}                     \\ \hline
\end{tabular}
\end{table}

\section{FUTURE WORK}

As future work, other than implementing other well-performing rules, which requires human expertise, it would be good to have a language of primitives from which the rules themselves could be evolved. Still in the topic of evolution, rather than have a list of rules that must be examined in order, a Neural Network with evolved weights could determine which rule (or action) to take, similarly to the approach of ~\cite{holmgaard2014evolving}. Alternatively, such controller could be developed by techniques such as Deep Reinforcement Learning~\cite{mnih2013playing}.

Another important development would be the ability to generate a model of the other players' behavior, such as how risk-taking they are when playing a card, their preferred discard policy, etc. If we could accurately recognize these features during game play, we could evolve specific chromosomes for playing with agents with those characteristics and so improve our mixed score. We are particularly interested in initiatives such as ~\cite{barrett2011empirical}, which attempts to build a model of unknown partners during cooperative gameplay by interpolating between known models. These models can also be used for non-rule-based agents, such as  MCTS~\cite{browne2012survey}, which require a model of the other players.

\section{CONCLUSION}

We used evolution in three steps to get better Hanabi-playing agents than the human-created baselines: First we evolved the order in which rules are applied, using a set of rules very similar to the ones used in ~\cite{walton2017evaluating}. As the quality of an agent depends not only on the ordering of the rules, but also on the expressiveness of the rule set, we then added rules that account for our partner's intentions (assuming a hinted card has high probability of being playable) and to choose which piece of information to give about a playable card in the least ambiguous way possible. This not only brought a quantitative increase to our score, but we also noticed qualitatively that the new rules were in general very effective, appearing at the head of many of the most successful chromosomes.

Finally, we created specialized agents for specific game sizes and using their behavior for any game size, we get an improvement over a generic agent that is optimized for playing all game sizes. This shows that the best strategy for Hanabi likely depends on the number of player. We analyzed 30 of our best chromosomes to attempt to identify patterns that make some strategies better in each game size, and also for mirror or mixed evaluation.

By combining evolution, new rules and specialized behavior, we get a improvement over the best purely rule-based agents, going from 18.16 to 19.32. While hat agents~\cite{cox2015make,bouzy2017playing} score significantly better than our mirror agents, they are unsuited for mixed or human play. To our knowledge, the only published non-hat agent that exceeds our score is the combination of Tree Search with a rule-based agent as evaluator, seen in ~\cite{bouzy2017playing}, with a score of 20.22 across all game sizes. However, that agent vastly exceeds the time budget allowed in the competition. It is also worth noting that, as future work, we could attempt to combine our own rule-based agent with Tree Search algorithms, or even use rule-based agents evolved specifically for this purpose.

In mixed mode, our improvements were smaller, but still expressive, going from 11.18 to 11.65. It is important to note that the test pool used for the mixed evaluation consists on some agents, such as Flawed and Random that are purposely very bad (scoring around 5 points in average), so scores in this mode are naturally expected to be lower. While we do not know which agents will be used as the competition test pool, we hope our improvements when pairing with the test pool of ~\cite{walton2017evaluating} translates to good results in the mixed competition.

\section*{ACKNOWLEDGMENT}

Rodrigo Canaan gratefully acknowledges the financial support from Honda Research Institute Europe (HRI-EU).

\addtolength{\textheight}{-12cm}  

\bibliographystyle{IEEEtran}
\bibliography{bibfile}

\begin{table*}[]
\centering
\caption{Appendix - Rule Set}
\label{table:ruleset}
\begin{tabular}{|l|p{8cm}|l|}
\hline
\textbf{Rule name}                                   & \textbf{Description}                                                                                                                     & \textbf{Category}                   \\ \hline
Play If Certain                                      & Plays a card with fully known information that is playable                                                                               &      1                          \\ \hline
Play Safe Card                                       & Plays a card that is known to be playable (even with partial information)                                                                &      1                          \\ \hline
Play probably safe card (p)                 & Plays card most likely to be playable if the probability of being playable is greater than p                                           & 1 $^{1}$    \\ \hline
If lives \textgreater{}1 play probably safe card (p) & Plays card most likely to be playable if the probability of being playable is greater than p and there is more than one available life & 2 $^{2}$         \\ \hline
If Hail Mary play probably safe card (p)             & Similar to above, but also requires the deck to be empty                                                                                 & 2 $^{3}$                        \\ \hline
Complete Tell Useful Card                            & Tells the missing information (color or value) of a partially known playable card to a player                                            &    1                            \\ \hline
Tell About Ones                                      & Tells a player about all their cards with value of one                                                                                   &    1                            \\ \hline
Tell Anyone About Useful Card                        & Tells a player some new information about one of their playable cards, prioritizing value if card is completely unknown.                 &   1                             \\ \hline
Tell Anyone About Oldest Useful Card                 & Tells a player some new information about their oldest playable card, prioritizing value.                                                &   1                             \\ \hline
Tell Playable Card                                   & Tells a player some information about a playable card. Decides randomly between value and color, even if the card is partially known     &    1                            \\ \hline
Tell Playable Card Outer                             & Same as Tell Anyone About Useful Card                                                                                                    &     1                           \\ \hline
Tell Anyone About Useless Card                       & Tells a player some information about a card that is never going to be playable                                                          &      1                          \\ \hline
Tell Dispensable                                     & Same as Tell Anyone About Useless Card                                                                                                   &       1                         \\ \hline
Tell Fives                                           & Tells a player about all their cards with value five                                                                                     &      1                          \\ \hline
Tell Most Information                                & Gives a hint that tells the most information, or most new information about their hand                                                   & 1 $^{4}$ \\ \hline
Tell Randomly                                        & Gives a random hint to a player                                                                                                          &   1                             \\ \hline
Tell Unknown                                         & Gives new information about a card in a player's hand.                                                                                   &    1                            \\ \hline
Discard Useless                                      & Discards cards whose pre-requisites have been discarded.                                                                                 &    1                            \\ \hline
Discard Safe                                         & Discards that is no longer playable.                                                                                                     &       1                         \\ \hline
Osawa Discard                                        & Discards a card that is useless or safe.                                                                                                 &      1                          \\ \hline
Discard If Certain                                   & Discards a card that with fully known information and no longer playable                                                                 &     1                           \\ \hline
Discard Highest                                      & Discards card in hand with highest known value.                                                                                          &     1                           \\ \hline
Discard Oldest                                       & Discards oldest card in hand.                                                                                                            &       1                         \\ \hline
Discard Oldet No Info First                          & Discards oldest card with no known information                                                                                           &      1                          \\ \hline
Discard Unidentified Card                            & Discards a card with no known information                                                                                                &    1                            \\ \hline
Discard least likely to be necessary                 & Discards card with smallest probability of being necessary for a perfect score                                                           &   1                             \\ \hline
Discard probably useless (p)               & Discards card with highest probability of being useless, as long as that probability is greater than p                                   & 1 $^{2}$      \\ \hline
Play Finesse                                         & Part of the Finesse strategy.                                                                                                            &    3                            \\ \hline
Play Finesse Told                                    & Part of the Finesse strategy.                                                                                                            &     3                           \\ \hline
Tell Finesse                                         & Part of the Finesse strategy.                                                                                                            &      3                          \\ \hline
Play Unique Possible Card                            & Part of the Finesse strategy                                                                                                             &     3                           \\ \hline
Tell Illinformed                                     & If a player is ill-informed, give them a hint.                                                                                           &      3                          \\ \hline
Try To Unblock                                       & If there is no unblocking player between you and a blocked one, unblock.                                                                 &     3                           \\ \hline
Legal Random                                         & Selects a random legal action                                                                                                            &     1                           \\ \hline
PlayJustHinted                                       & See section \ref{section:experiments}                                                                                                            &     4                           \\ \hline
TellUnambiguous1                                       & See section \ref{section:experiments}                                                                                                            &     4                           \\ \hline
TellUnambiguous2                                       & See section \ref{section:experiments}                                                                                                            &     4                           \\ \hline
\end{tabular}
\centering{\\\textbf{Categories}: \\
1: Already implemented in the framework and similar to a rule described in ~\cite{walton2017evaluating}\\
2: Implemented using the framework's IF and/or CONDITIONAL rule and described in ~\cite{walton2017evaluating}\\
3: Implemented in the framework, but not described in ~\cite{walton2017evaluating} \\
4: New rules implemented for this work \\
\textbf{Parameters:} \\
$^{1}$ - Values of p = 0, 0.2, 0.25, 0.4, 0.6 and 0.8 were used for this rule\\
$^{2}$ - Values of p = 0, 0.2, 0.4, 0.6 and 0.8 were used for this rule\\
$^{3}$ - Values of 0 and 0.1 were used for this rule \\
$^{4}$ - Only the variant that tells the most information was used for this rule}
\end{table*}

\end{document}